\documentclass{article}
\usepackage{ijcai19}

\usepackage{amsmath}
\usepackage{algorithm, algpseudocode}
\usepackage{mathrsfs}
\usepackage{titlesec}

\usepackage{times}
\usepackage{soul}
\usepackage{url}
\usepackage[hidelinks]{hyperref}
\usepackage[utf8]{inputenc}
\usepackage[small]{caption}
\usepackage{booktabs}
\usepackage{graphicx}
\usepackage{amsfonts}
\usepackage{color}
\usepackage{xcolor}
\usepackage{enumitem}

\title{Spectral Autoencoder for Anomaly Detection in Attributed Networks}

\author{Yuening Li\dag, Xiao Huang\dag, Jundong Li\ddag, Mengnan Du\dag, Na Zou\dag\\
\affiliations
\dag Texas A\&M University, College Station, USA\\
\ddag University of Virginia, Charlottesville,  USA  \\
\emails
\{yueningl,xhuang,dumengnan,nzou1\}@tamu.edu, jl6qk@virginia.edu 
}

\begin{document}

\maketitle

\begin{abstract}
    Anomaly detection aims to distinguish observations that are rare and different from the majority. While most existing algorithms assume that instances are i.i.d., in many practical scenarios, links describing instance-to-instance dependencies and interactions are available. Such systems are called attributed networks. Anomaly detection in attributed networks has various applications such as monitoring suspicious accounts in social media and financial fraud in transaction networks. However, it remains a challenging task since the definition of anomaly becomes more complicated and topological structures are heterogeneous with nodal attributes. In this paper, we propose a spectral convolution and deconvolution based framework - SpecAE, to project the attributed network into a tailored space to detect global and community anomalies. SpecAE leverages Laplacian sharpening to amplify the distances between representations of anomalies and the ones of the majority. The learned representations along with reconstruction errors are combined with a density estimation model to perform the detection. They are trained jointly as an end-to-end framework. Experiments on real-world datasets demonstrate the effectiveness of SpecAE.

\end{abstract}

\section{Introduction}
Anomaly detection targets at identifying the rare instances that behave significantly different from the majority instances. Conventional detection algorithms are mainly based on the assumption that instances are independent and identically distributed (i.i.d.)~\cite{ruff2018deep}. But in many practical scenarios, instances are often explicitly or implicitly connected with each other,  resulting in the so-called \emph{attributed networks}~\cite{huang2019graph}. Attributed networks are increasingly used to represent various real-world systems since the synergy between network structures and nodal attributes. For instance, in social media, users are not only affiliated with profile information and personalized posts as nodal attributes, but also linked with each other by different social relations~\cite{liu2017accelerated}. Detecting anomalies in attributed networks have witnessed a surge of research interests from various domains, such as suspicious account detection in social media, abuse monitoring in healthcare systems, financial fraud monitoring, and network intrusion detection in cyber security~\cite{song2018deepmem}.


 
 However, the unique data characteristics of attributed networks bring several challenges to anomaly detection. Firstly, the definition of anomaly becomes more complicated and obscure. Apart from anomalous nodes whose nodal attributes are rather different from the majority reference nodes from a global perspective, nodes with nodal attributes deviate remarkably from their communities are also considered to be anomalies~\cite{li2017radar}. Secondly, the topological structures are within irregular or non-Euclidean spaces~\cite{Defferrard-etal16Convolutional}, and traditional anomaly detection methods~\cite{hawkins2002outlier} could not be directly applied to attributed networks. Thirdly, networks and nodal attributes are heterogeneous, and effective anomaly detection methods need to fuse these two data modalities more synergistically.
 
To handle the topological structures and the heterogeneity challenge, in network analysis, a widely-used and effective approach is to embed all information in attributed network into unified low-dimensional node representations. However, it would achieve suboptimal performance when directly applied to the anomaly detection. To achieve the join embedding, most existing network embedding algorithms~\cite{liu2019single} and recent joint embedding based anomaly detection methods~\cite{li2019deepstruc} mainly rely on the homophily hypothesis, which implies that connected nodes tend to have similar nodal attributes~\cite{mcpherson2001birds}. Based on this assumption, topological relations could be introduced via involving regularization terms to minimize the distance between embedding representations of linked nodes~\cite{yu2018netwalk}. Another effective way to encode networks is to employ graph convolutional networks (GCN)~\cite{kipf2016semi}. It could be considered as a special form of Laplacian smoothing that learns nodes' representations from their neighbors~\cite{li2018deeper}. The homophily hypothesis or smoothing operations are not in line with anomaly detection. They might over-smooth the node representations, and make anomalous nodes less distinguishable from the majority within the community. For example, malicious users would have completely different nodal attributes than their friends. Customers who have written fake reviews might purchase the same products as the normal ones. Thus, existing joint learning models in network analysis could not be directly applied to anomaly detection.

To bridge the gap, we investigate the problem of anomaly detection in attributed networks. Two specific research questions are studied. 1) How to formally define anomalies in attributed network? 2) How to perform joint learning on topological structures and nodal attributes, while being in line with the anomaly detection, such as getting rid of the restriction of homophily and over-smooth problems? Following these research perspectives, our contributions could be summarized as follows.


\begin{itemize}
\item We propose a graph convolution and deconvolution based framework - SpecAE, for anomaly detection in attributed networks.

\item We develop a tailored model to project the attributed network into a special space for global anomalies and community anomalies. It leverages Laplacian sharpening to amplify the distances between embedding representations of anomalies and the ones of other nodes within the communities.


\item We conduct evaluations on real-world datasets, demonstrating the effectiveness of SpecAE in terms of anomaly detection in attributed networks.
\end{itemize}

\begin{figure*}
\centering
\includegraphics[height=2.4
in]{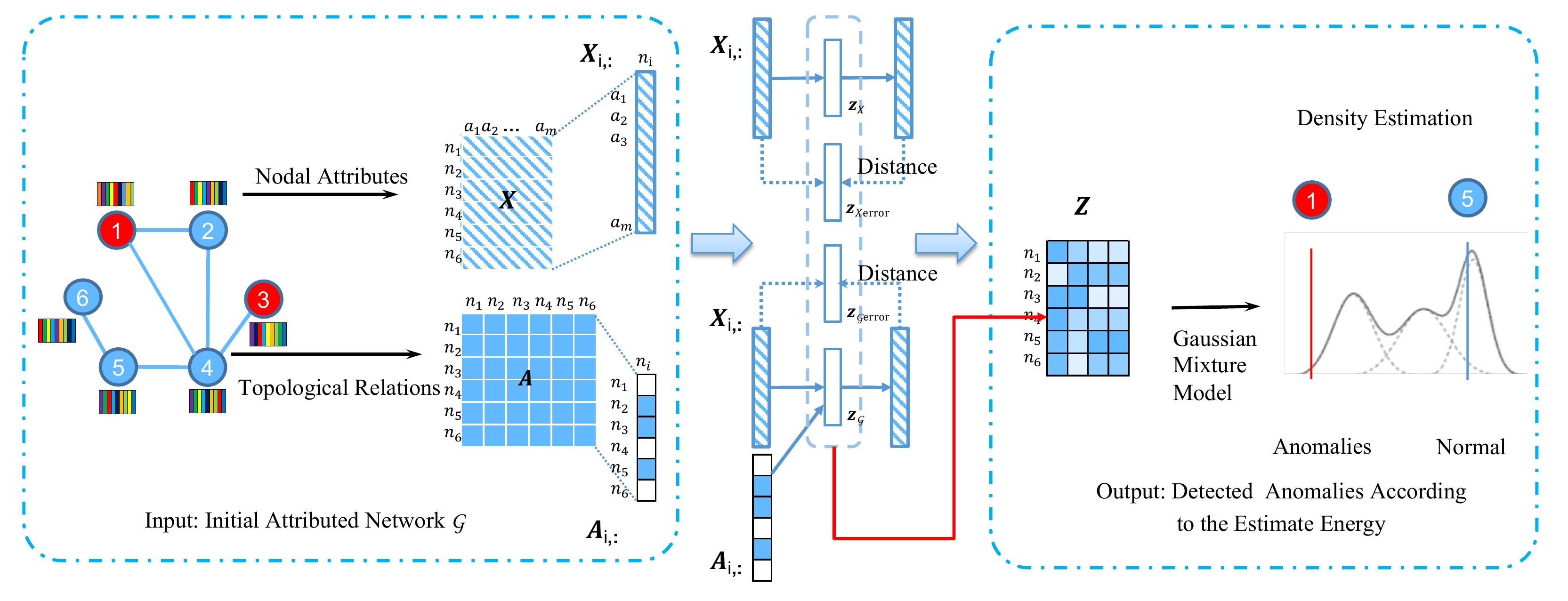}
\vspace{-8pt}
\caption{A spectral autoencoder based anomaly detection approach for attributed networks framework.}
\label{Structure}
\vspace{-3pt}
\end{figure*}
\vspace{-8pt}
\section{Problem Statement}
We use bold upper case letters (e.g., $\mathbf{X}$) to denote matrices, bold lowercase characters for vectors (e.g., $\mathbf{x}$). We denote the $a^{th}$ row of the matrix $\mathbf{X}$ as  $\mathbf{X}_{a,:}$, the $b^{th}$ entry of the vector $\mathbf{x}$ as $\mathbf{x}_{b}$.  Let $\mathcal{G}$ = \{$\mathbf{A}$, $\mathbf{X}$\} be an input attributed network, with $n$ nodes interconnected by a network. $\mathbf{A} \in \mathbb{R}^{n \times n}$ denotes the corresponding adjacency matrix. Each node is associated with a set of $m$-dimensional nodal attributes. We collect all these nodal attribute as $\mathbf{X}\in \mathbb{R}^{n \times m}$. 

{\bf Definition 1 (Global Anomaly)} {\em It refers to a node whose nodal attributes are rare and significantly different from the nodal attributes of majority nodes, from a global perspective.}

{\bf Definition 2 (Community Anomaly)} {\em It is defined as a node $i$ with nodal attributes that significantly deviate from the node $i$'s neighbors' nodal attributes, based on the topological network structure.
}

We define two types of anomalies. Detecting them in attributed networks is crucial. For instance, in social networks, the malicious, zombie, spam, or suspicious users' behaviors might deviate with most of the normal users'. We treat this type of users as global anomalies. There also could be a small portion of accounts with contents significantly different from their community, such as advertising/marketing users in college student clubs. We treat this type of users as community anomalies.

Given the aforementioned definitions, we formally define the anomaly detection in attributed networks problem as follows. Given a set of nodes $\mathcal{V}$ connected by an attribute network $\mathcal{G}$, we target at identifying the global anomalies and community anomalies in $\mathcal{V}$.

\section{Anomaly Detection in Attributed Networks}
We propose a Spectral autoencoder based anomaly detection framework - SpecAE, for attributed networks. The pipeline of SpecAE is illustrated in Fig.1. Given an attributed network $\mathcal{G}$, we leverage a tailored Spectral autoencoder to jointly embed nodal attributes and relations into a carefully-designed space $\mathbf{Z} =[\mathbf{Z}_{X}; \mathbf{Z}_{X\text{error}}; \mathbf{Z}_{\mathcal{G}};  \mathbf{Z}_{\mathcal{G}\text{error}}]$, which consists of four sources of components. To detect the global anomalies, we apply an autoencoder to all nodal attributes ${\bf X}$ to learn embedding representations ${\bf Z}_X$ as well as the reconstruction errors $\mathbf{Z}_{X\text{error}}$. In such a way, we could globally compare all nodes' attributes. To detect the community anomalies, we design novel graph convolutional encoder and deconvolutional decoder networks to learn nodes' community representations $\bf{Z}_{\mathcal{G}}$, based on each node's neighbors. The corresponding reconstruction errors are denoted as $\mathbf{Z}_{\mathcal{G}\text{error}}$. Later on, based on the tailored joint representations ${\bf Z}$, we estimate the suspicious level of each node $u$ by measuring its embedding representation ${\bf Z}_{i,:}$'s energy in the Gaussian Mixture Model.

\subsection{Tailored Embedding Spaces for Global Anomalies and Community Anomalies}
Attributed networks bring both opportunities and challenges to the anomaly detection task. The two sources $\bf A$ and $\bf X$ bring more information than uni-modality, but make the definition of anomaly more complex.

To perform anomaly detection in $\mathcal{G}$, an intuitive approach is to apply the attributed network embedding~\cite{yu2018netwalk} algorithm to project $\mathcal{G}$ into unified node embedding representations, and estimate the suspicious levels of nodes based on the learned representations. The goal of network embedding is to preserve nodes' major characteristics and map all nodes into a unified space, based on the homophily hypothesis~\cite{mcpherson2001birds} that nodes with similar topological structures tend to have similar node attributes. However, anomaly detection focuses on discriminating nodes with characteristics that are different from the majority, including nodes that are inconsistent with the homophily hypothesis. Thus, the conventional attributed network embedding methods will lead to a suboptimal result.

 In this paper, we propose to category anomalies in attributed networks into two classes, i.e., global anomalies and community anomalies. To distinguish these anomalies, we project $\mathcal{G}$ into two types of tailored low-dimensional spaces. 

First, based on the definition of global anomaly, we employ an autoencoder to embed nodal attributes $\bf X$ to learn the first type of tailored representations, i.e., $\mathbf{Z}_{X}$ and $\mathbf{Z}_{X\text{error}}$. The encoding and decoding processes are achieved via: 
\begin{equation}
\begin{aligned}
\mathbf{Z}_{X} = \sigma \left (\mathbf{b}_{e}+ \mathbf{X}\mathbf{W}_{e}\right), \\ \hat{\bf X} = \sigma \left (\mathbf{b}_{d}+ \mathbf{Z}_{X}\mathbf{W}_{d}\right), 
\label{z1}
\end{aligned}
\end{equation}
where $\hat{\bf X}$ denotes the reconstructed nodal attributes. Since anomalies are harder to reconstruct than normal nodes~\cite{chen2017outlier}, we also include the reconstruction errors $\mathbf{Z}_{X\text{error}}$.
\begin{equation}
\mathbf{Z}_{X\text{error}}={\text{dis}} (\mathbf{X}, \hat{\bf X}), 
\label{z2}
\end{equation}
where the operation $\text{dis}(\cdot,\cdot)$ denotes a series of the distance measuring matrices such as the Euclidean distance and the cosine distance. In such a way, global anomalies would tend to have their representations in $\mathbf{Z}_{X}$ being different from the majority, as well as $\mathbf{Z}_{X\text{error}}$ being large.

Second, to identify community anomalies, we need to jointly consider $\bf A$ and $\bf X$. Based on their topological dependencies, we develop a novel graph convolutional  encoder and graph deconvolutional  decoder to learn the second type of tailored representations, i.e., $\mathbf{Z}_{\mathcal{G}}$ and $\mathbf{Z}_{\mathcal{G}\text{error}}$. Details are introduced as follows.


\subsection{Graph Convolution and Deconvolution}
Our goal is to learn nodes' community representations $\mathbf{Z}_{\mathcal{G}}$, which describe the expected behaviors of nodes according to their neighbors.

A straightforward solution is to apply GCN to embed $\mathcal{G}$. From a formulation perspective~\cite{li2018deeper}, GCN could be treated as a special form of \textit{Laplacian Smoothing}. 
The \textit{Laplacian smoothing}  on each channel of the input features can be written as:
  \begin{equation}
    \label{laplacianlayer}
  \mathbf{Y} =  (\mathbf{I}-\gamma\mathbf{\tilde{D}}^{-1}\mathbf{(\mathbf{\tilde{D}}-\mathbf{\tilde{A}})}) \mathbf{X},
  \end{equation}
where $0 < \gamma \leq1$ is a parameter which controls the weighting
between the features of the current node and the features of its neighbors, with a normalization trick as:
\begin{equation}
  \mathbf{I} + \mathbf{D}^{-\frac{1}{2}} \mathbf{A}\mathbf{D}^{-\frac{1}{2}} \rightarrow \mathbf{\tilde{D}}^{-\frac{1}{2}} \mathbf{\tilde{A}}\mathbf{\tilde{D}}^{-\frac{1}{2}}, 
\end{equation}
where $\mathbf{\tilde{A}}: = \mathbf{A}+\mathbf{I}$ and $\mathbf{\tilde{D}}: = \sum_{j}\mathbf{\tilde{A}}_{ij}$.

Based on the definition of convolution in graph signals on the spectral domain, to generate a new matrix $\mathbf{Y}$ from $\mathbf{X}$ by applying the graph convolution as spectral filters, we have:
  \begin{equation}
  \label{gcnlayer1}
  \mathbf{Y} =  \mathbf{\tilde{D}}^{-\frac{1}{2}} \mathbf{\tilde{A}}\mathbf{\tilde{D}}^{-\frac{1}{2}} \mathbf{X}, 
  \end{equation}
which is a special form of \textit{Laplacian Smoothing} when $\gamma = 1$ after replacing the \textit{normalized graph Laplacian} $\mathbf{L:=\tilde{D}}^{-1}\mathbf{\tilde{A}}$ with the \textit{symmetric normalized graph Laplacian} $\mathbf{L:=\tilde{D}}^{-\frac{1}{2}} \mathbf{\tilde{A}}\mathbf{\tilde{D}}^{-\frac{1}{2}} $.

However, GCN does not contain nodal attributes reconstruction procedure, which is useful in the anomaly detection task with three reasons.
i) Training transformation functions base on one-class observations (only normal instances available) without the reconstruction error may easily lead the objective function converges to a local optimal when all of the node representations collapse into a small area  (one point in extreme cases) in the latent space. 
ii) Repeatedly applying \textit{Laplacian smoothing} might cause the nodal attributes over-mixed with their neighbors and make them indistinguishable, since it will be more difficult to identify each individual instance after the smoothness operation which has weakened their uniqueness of the original attributes. iii) Reconstruction error usually contains useful information as indicators for anomaly detection. For example, in ~\cite{hawkins2002outlier}~\cite{sakurada2014anomaly}~\cite{an2015variational}~\cite{chen2017outlier}, the reconstruction error could be directly used as an anomaly score to rank the anomalous degree of nodes.
   
Thus, we decide to design the graph decoding (graph deconvolution) from the smoothed features as a complementary inverse process of graph convolution. Inspired by the digital images field, \textit{sharpening} is an inverse process of blurring/smoothing~\cite{ma2014optimized}. Comparing with the smoothing process, which is done in the spatial domain by pixel averaging in neighbors, sharpening is to find the difference by the neighbors, done by spatial differentiation. A Laplacian operator is to restore fine details of an image which has been smoothed or blurred. After Laplacian sharpening, we can reconstruct the nodal attributes from the fusion features caused by the graph convolution process.

If we replace the original GCN function in Eq.~\eqref{gcnlayer1} as a general \textit{Laplacian smoothing}, we will have:
  \begin{equation}
      \begin{aligned}
  \mathbf{Y}= \mathbf{X} - \alpha (\mathbf{X} -  \mathbf{\tilde{D}}^{-\frac{1}{2}} \mathbf{\tilde{A}}\mathbf{\tilde{D}}^{-\frac{1}{2}} \mathbf{X}) \\
   =   (1-\alpha)\mathbf{X} + \alpha \mathbf{\tilde{D}}^{-\frac{1}{2}} \mathbf{\tilde{A}}\mathbf{\tilde{D}}^{-\frac{1}{2}} \mathbf{X}.
      \end{aligned}
  \end{equation}
Generalizing the above definition of graph convolution with adjacency matrix $\mathbf{A} \in \mathbb{R}^{n \times n}$  and the nodal attributes $\mathbf{X} \in \mathbb{R}^{n \times m}$ of $n$ instances and $m$ input channels, the propagation rule of the convolution layer can be written as:
    \begin{equation}
    \label{lapsmooth}
  \text{Conv} (\mathbf{X, A}) = \sigma \left ( (1-\alpha)\mathbf{X} + \alpha \mathbf{\tilde{D}}^{-\frac{1}{2}} \mathbf{\tilde{A}}\mathbf{\tilde{D}}^{-\frac{1}{2}} \mathbf{X}\right)\mathbf{W}_{f} , 
  \end{equation}
  where $\mathbf{W}_{f}$ is the trainable weight matrix in the convolution layer, $\sigma$ is the activation function, e.g., $ReLU ( \cdot )$. The parameter $\alpha$ is to control the weighting between the features of the current nodal attributes and the features of its neighbors. When  $\alpha = 1$, we can treat the Eq.~\eqref{lapsmooth} as a FC layer in a MLP framework; if $\alpha = 0$, we have one graph convolutional layer. We conduct experiments under different parameter settings in the ablation studies section to illuminate the effect of the $\alpha$.

Contrary to \textit{Laplacian smoothing}, we compute the new features of nodal attributes through sharpening the features with their neighbors in order to reconstruct the features from the smoothed results.
To magnify the difference between the current node and its neighbors, we will have:
  \begin{equation}
      \begin{aligned}
  \mathbf{Y}= \mathbf{X} + \alpha (\mathbf{X} -  \mathbf{\tilde{D}}^{-\frac{1}{2}} \mathbf{\tilde{A}}\mathbf{\tilde{D}}^{-\frac{1}{2}} \mathbf{X}) \\
   =   (1+\alpha)\mathbf{X} - \alpha \mathbf{\tilde{D}}^{-\frac{1}{2}} \mathbf{\tilde{A}}\mathbf{\tilde{D}}^{-\frac{1}{2}} \mathbf{X}. 
      \end{aligned}
    \label{fxa}
  \end{equation}
Given the above definition of deconvolution to an attributed network $\mathcal{G}$, with adjacency matrix $\mathbf{A} \in \mathbb{R}^{n \times n}$  and the nodal attributes $\mathbf{Z} \in \mathbb{R}^{n \times m}$ of $n$ instances and $m$ input channels which after the convolution process, the propagation rule of the deconvolution layer is:
     \begin{equation}
  \mathrm{Deconv} (\mathbf{Z, A}) = \sigma \left ( (1+\alpha)\mathbf{Z} - \alpha \mathbf{\tilde{D}}^{-\frac{1}{2}} \mathbf{\tilde{A}}\mathbf{\tilde{D}}^{-\frac{1}{2}} \mathbf{Z}\right)\mathbf{W}_{g}, 
    \label{gza}
  \end{equation}
where $\mathbf{W}_{g}$ is the trainable weight matrix in the deconvolution layer. 
After the sharpening process, we can reconstruct the original attributes from the smoothed features.

\begin{algorithm}[t]
\label{pseudoPSO}
\begin{algorithmic}[1]
\State \textbf{Input:} Initialize  a target attribute network $\mathcal{G}$ with the nodal attributes $\mathbf{X}$ $\in$ $\mathbb{R}^{n \times m}$ and network topology relations $\mathbf{A}$ $\in$ $\mathbb{R}^{n \times n}$, with hyper-parameters $\lambda_{1}$, $\lambda_{2}$, $\alpha$
\State \textbf{Output:} A node list $l$ sorted with normality score

\State Initialize $\mathbf{W}$, $\mathbf{\hat{\phi}}$, $\mathbf{\hat{\mu}}$, $\mathbf{\hat{\Sigma}}$, and segment $\mathbf{X}$ and $\mathbf{A}$
\State Randomly select $k$ samples with normal behavior out of $n$ instances as training samples 
\For{each epoch i in parallel } 
    \State $\mathbf{W}^{i+1}$ $\leftarrow$ $\mathbf{W}^{i}$ - ${\nabla}_{\mathbf{W}} J (\mathbf{W}$) in Eq.~\eqref{objective}
    \State Update $\mathbf{\hat{\phi}}$ in Eq.~\eqref{phi}
    \State Update $\mathbf{\hat{\mu}}$ in Eq.~\eqref{mu}
    \State Update $\mathbf{\hat{\Sigma}}$ in Eq.~\eqref{Sigma} 
\EndFor
\State {Compute sample energy of all n samples according to Eq.~\eqref{energy} as the normality score for each n $\in$ $[1, n]$}
\State {Return the sorted list of nodes $l$  $\in$ $[1, n-k]$ which decrease sorted by the normality score}
\end{algorithmic}\label{ag1}
\caption{SpecAE Framework for Anomaly Detection}
\end{algorithm}

Given nodal attributes $\mathbf{X}$ with adjacency matrix $\mathbf{A}$ , the compression network computes their low-dimensional representations $\mathbf{Z}$ as follows:
\vspace{-8pt}
\begin{equation}
\begin{aligned}
&f (\mathbf{Z}_{\mathcal{G}}|\mathbf{X, A}) = \prod _{i=1}^{n}f (\mathbf{Z}_{\mathcal{G}_{i,:}}|\mathbf{X}_{i,:}, \mathbf{A}), \\
\vspace{-4pt}
 \mathrm{with} & \quad f (\mathbf{Z}_{\mathcal{G}_{i,:}}|\mathbf{X, A})  = \mathcal{N}  (\mathbf{Z}_{\mathcal{G}_{i,:}}|\mathbf{\mu}_{i,:}, \text{diag} (\mathbf{\sigma}_{i,:}^2));
\label{z3}
\end{aligned}
\end{equation}
\vspace{-4pt}
\begin{equation}
\begin{aligned}
&g (\widetilde{\mathbf{X}}|\mathbf{Z}_{\mathcal{G}}, \mathbf{A}) = \prod _{i=1}^{n}g (\widetilde{\mathbf{X}}_{i,:}|\mathbf{Z}_{\mathcal{G}_{i,:}}, \mathbf{A}), \\
\vspace{-4pt}
 \mathrm{with} & \quad g (\widetilde{\mathbf{X}}_{i,:}|\mathbf{Z}_{\mathcal{G}_{i,:}},\mathbf{A})  = \text{Deconv}(\mathbf{Z}_{\mathcal{G}_{i,:}},\mathbf{A});
\end{aligned}
\end{equation}
\begin{equation}
\mathbf{Z}_{\mathcal{G}\text{error}}={\text{dis}} (\mathbf{X}, \widetilde{\mathbf{X}}), 
\label{z4}
\end{equation}
    \begin{equation}
  \mathbf{Z} = [\mathbf{Z}_{X}; \mathbf{Z}_{\mathcal{G}}; 
  \mathbf{Z}_{X\text{error}}; \mathbf{Z}_{\mathcal{G}\text{error}}],
  \label{z5}
  \end{equation}
  where the $\mathbf{\mu}_{i,:} := \text{Conv}_{\mathbf{\mu}}(\mathbf{X},\mathbf{A})$ denote the mean vector in Eq.~\eqref{fxa}, similarly, $\text{log}\mathbf{\sigma}_{i,:} := \text{Conv}_{\mathbf{\sigma}}(\mathbf{X},\mathbf{A})$. The latent variable $\mathbf{Z}$ of each node is shown in Eq.~\eqref{z5}, which concatenates from four components in Eq.~\eqref{z1}, Eq.~\eqref{z2}, Eq.~\eqref{z3}, and Eq.~\eqref{z4}. 
  $\mathbf{Z}_{\mathcal{G}}$ is learned by the graph convolutional encoder which including the neighbor attribution with the topological relations. $\mathbf{Z}_{\mathcal{G}\text{error}}$ is the reconstruction of the nodal attributes $\mathbf{X}$ base on the fusion representation $\mathbf{Z}_{\mathcal{G}}$ and the adjacency matrix  $\mathbf{A}$. 

\subsection{Anomaly Detection via Density Estimation}
Given the learned embedding representation $\mathbf{Z}$ of $N$ samples and their soft mixture-component membership prediction $\hat{\gamma}$  (estimated from a softmax layer based on the low-dimensional representation $\mathbf{Z}$), where $\hat{\gamma}$ is a K-dimensional vector and $K$ is the number of mixture components in Gaussian Mixture Model (GMM), we can estimate the mixture probability $\hat{\phi}_{k}$, the mean value $\hat{\mu}_{k}$, the covariance vector $\hat{\Sigma}_{k}$ for component $k$ in GMM respectively~\cite{zong2018deep}.
 
The sample energy can be inferred by:
\vspace{-8pt}
    \begin{equation}
   E (\mathbf{z})=-\log\left (\sum_{k=1}^{K}\hat{\phi}_{k}\frac{\exp -\frac{1}{2} (\mathbf{z}-\hat{\mu}_{k})^{T}\hat{\Sigma}_{k}^{-1} (\mathbf{z}-\hat{\mu}_{k})}{\sqrt[]{|2\pi\hat{\Sigma}_{k}|}}  \right), 
   \label{energy}
   \end{equation} 

where we estimate the parameters in GMM as follows:
 \begin{equation}
	\mathbf{\hat{\gamma}} = {\text{softmax}} (\mathbf{z}), 
  \end{equation} 
\vspace{-6pt}
  \begin{equation}
  	\hat{\phi}_{k} = \sum_{i=1}^{N}\frac{\hat{\gamma}_{ik}}{N}, 
  	\label{phi}
  \end{equation} 
\vspace{-4pt}
  \begin{equation}
  	\hat{\mu}_{k} = \frac{\sum_{i=1}^{N}\hat{\gamma}_{ik}\mathbf{z}_{i}}{\sum_{i=1}^{N}\hat{\gamma}_{ik}}, 
  	\label{mu}
  \end{equation}
\vspace{-4pt}
  \begin{equation}
  	\mathbf{\hat{\Sigma}}_{k} = \frac{\sum_{i=1}^{N}\hat{\gamma}_{ik} (\mathbf{z}_{i}-\hat{\mu}_{k}) (\mathbf{z}_{i}-\hat{\mu}_{k})}{\sum_{i=1}^{N}\hat{\gamma}_{ik}}^{T}.
  	\label{Sigma}
  \end{equation}

\subsection{Objective Function}
Given an input attributed network $\mathcal{G}$ = $\mathbf{\{A, X\}}$, the objective function 
is constructed as follow:
  \begin{equation}
    \begin{aligned}
	J (\mathbf{W}) = & \mathbb{E}[{\text{dis}} (\mathbf{X}, \hat{\mathbf{X}})] +\mathbb{E}[{\text{dis}} (\mathbf{X}, \widetilde{\mathbf{X}})] + \lambda_{1}\mathbb{E}(E(\mathbf{\mathbf{Z}}))\\
	+ & \lambda_{2}\sum_{i=1}^{K}\sum_{j=1}^{d}\frac{1}{\mathbf{\hat{\Sigma}}_{ij}} - \mathcal{KL}[f (\mathbf{Z}_{\mathcal{G}}|\mathbf{X, A})||g (\mathbf{Z}_{\mathcal{G}})]\\
	+ & \mathbb{E}_{f (\mathbf{Z}_{\mathcal{G}}|\mathbf{X, A})} \log(g (\widetilde{\mathbf{X}}|\mathbf{Z}_{\mathcal{G}}, \mathbf{A})).\\
    \end{aligned}
    \label{objective}
  \end{equation} 
The objective function includes four components:
\begin{itemize}[leftmargin=*]

\item $\mathbb{E}[{\text{dis}} (\mathbf{X}, \hat{\mathbf{X}})] , \mathbb{E}[{\text{dis}} (\mathbf{X}, \widetilde{\mathbf{X}})]$ are the loss function that characterizes the reconstruction error.
\item $E (\mathbf{Z})$ denotes the sample energy of the GMM estimation in Eq.~\eqref{energy}. Through minimizing the sample energy, we maximize the likelihood of non-anomalous samples, and predict samples with top-K high energy as anomalies.
\item To avoid trivial solutions when the diagonal entries of covariance matrices degenerate to 0, we penalize small values by the third component as a regularizer.
\item We optimize the variational lower bound as the last two terms. $\mathcal{KL}[f (\cdot)||g (\cdot)]$ is the Kullback-Leibler divergence between $f (\cdot)$ and $g (\cdot)$. $g (\cdot)$ is a Gaussian prior $g (\mathbf{Z}_{\mathcal{G}})  = \prod _{i=1}^{n}g (\mathbf{Z}_{\mathcal{G}_{i,:}}) = \mathcal{N}  (\mathbf{Z}_{\mathcal{G}_{i,:}}|0, \mathbf{I})$.

\end{itemize}

After optimizing the objective function, our proposed  approach can be applied to detect anomalous to the input attributed network. For our testing data, we can rank the anomalous degree of each node according to the corresponding estimation energy in Eq.~\eqref{energy}. According to the ranked energy, nodes with larger scores are more likely to be considered as anomalies.

\section{Experiments}
In this section, we evaluate the performance of our model with experiments on real-world datasets, to verify the effectiveness of the proposed approach. 
\subsection{Datasets}

We adopt two real-world datasets to evaluate the effectiveness of SpecAE. The detailed descriptions are shown in Tab.~\ref{detailedTable}. Cora is a network consisting of $3$,$327$ scientific publications, with $5$,$429$ links to denote the citation relations. Each article is described by one 0/1-valued word vector indicating the absence/presence of the corresponding word from the dictionary as node attributes. Pubmed comprises $19$, $717$ publications with $44$,$338$ citations from biomedical field, with $500$ dimension vector descriptions as nodal attributes. We also conduct a case study on another dataset PolBlog~\cite{perozzi2014focused}, which contains various blog content mainly related to political debates written by different liberal bloggers during the Iraq war in 2005.

In order to simulate ground truth outlierness, for Cora and Pubmed, we adopt the same strategy as in~\cite{skillicorn2007detecting,song2007conditional} to generate a combined set of anomalies from nodal attributes perspective and topological structure perspective. We inject an equal ratio of anomalies for both perspectives. First, we randomly select $m$ bags of words from the word dictionary which have low correlations as new nodal attributes, and replace these generated nodal attributes to the original $\mathbf{X}$, mark them as anomalies; Then we randomly pick another $m$ nodes and replace the attribution of each node $i$ to another node $j$ where the node $j$ deviates from the node $i$ with different context attributes (e.g., in citation networks, node $i$ and $j$ denote different categories of paper), in the meanwhile we keep their original topological relations (we keep the adjacency matrix $\mathbf{A}$ to denote their citation relations).

\subsection{Baseline Methods}
\begin{table}
    \centering
    \small
\begin{tabular}{|l|cc|}
\hline
Dataset       & Cora  & Pubmed \\ \hline
\# nodes       & $3$,$327$ & $19$,$717$  \\
\# edges       & $5$,$429$ & $44$,$338$  \\
\# attributes  & $1$,$433$ & $500$   \\
anomaly ratio & $5$\%    & $5$\%   \\ \hline
\end{tabular}
\vspace{-4pt}
\caption{Statistics of Cora and Pubmed. }
\vspace{-6pt}
\label{detailedTable}
\end{table}
\begin{figure}[t]
\includegraphics[width=1.0\columnwidth]{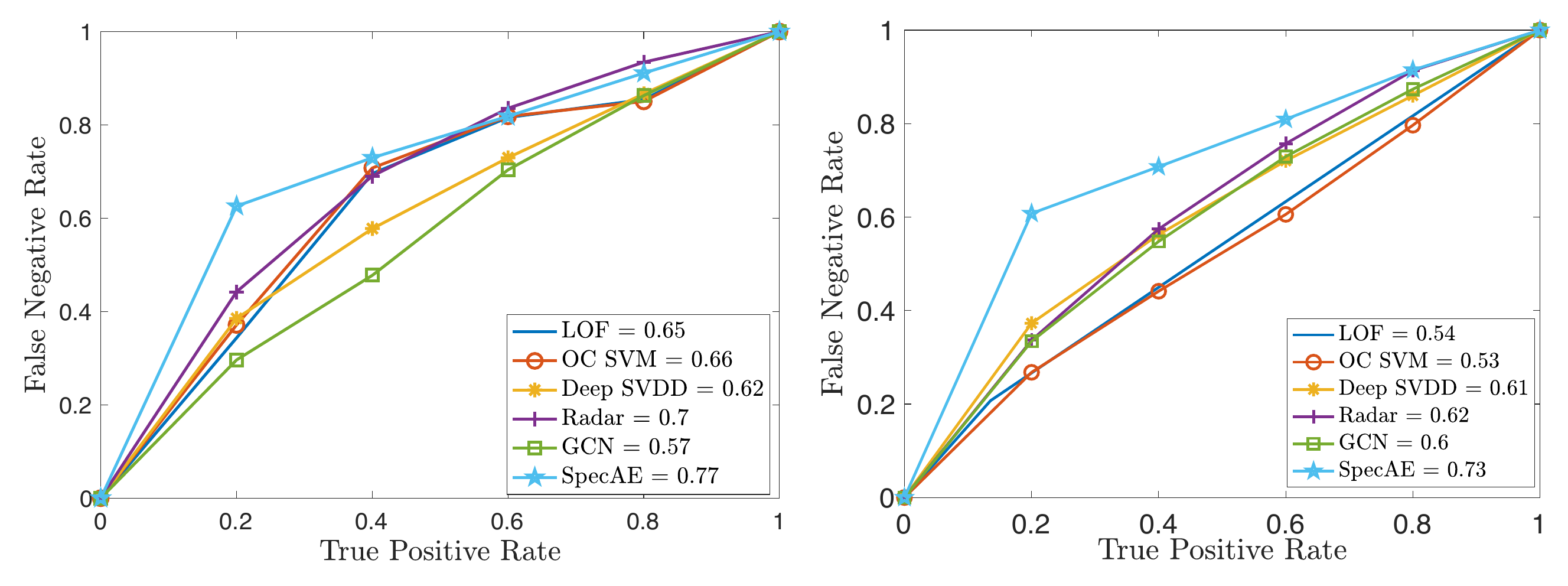}
\centering
\vspace{-20pt}
\caption{ROC and AUC on Cora and Pubmed (from left to right).}
\label{ROCCora}
\vspace{-8pt}
\end{figure}

We compare SpecAE with the five baseline methods:

\begin{itemize}
\item \textbf{LOF}~\cite{breunig2000lof}: Detects outliers out of the samples which have a substantially lower density. LOF based methods compute the local density deviation of nodal attributes with respect to their neighbors.
\item \textbf{OC-SVM}~\cite{scholkopf2000support}: Estimates the boundary of distribution that encompasses most of the observations (nodal attributes), and then labels the outliers which lie far from the learned boundary.
\item \textbf{Deep-SVDD}~\cite{ruff2018deep}: Trains a neural network while minimizing the volume of a hypersphere that encloses the network representations of the nodal attributes, analyzes the abnormality of each node based on the distance between the nodal attributes and the center of the learned sphere.
\item \textbf{Radar}~\cite{li2017radar}: Detects anomalies whose behaviors are different from the majority by characterizing attribute reconstruction
residuals and its correlation with topological relations and nodal attributes.
\item \textbf{GCN}~\cite{kipf2016semi}: 
Combines graph structures and nodal features in the convolution, and propagates over the graph through multiple layers. In our experiment, we modify the structure of the original GCN to accommodate the task of anomaly detection. We add the same density estimation layers as SpecAE after the graph convolutions for fair comparisons.
\end{itemize}

As shown in the introduction above, we compare SpecAE with three types of baselines. First, to investigate the effectiveness of utilizing features from both topological dependencies and nodal attributes instead of from single modality, we include LOF and OC-SVM. Second, to show the capacity of deep structures, we compare the performance between the shallow models (LOF, OC-SVM and Radar), as well as deep structured methods (Deep-SVDD and GCN). Third, to show the effectiveness of the proposed SpecAE, we modify the vanilla GCN for the specific anomaly detection task.

\subsection{Experimental Results}

\begin{table}[]
\tiny
\setlength{\tabcolsep}{5.5pt}
\centering
\begin{tabular}{|c|cccc|cccc|}
\hline
\multicolumn{9}{|c|}{\textbf{Accuracy@K}}                                                                                                                                                                       \\ \hline
\textbf{}                                                                                      & \multicolumn{4}{c|}{\textbf{Cora}} & \multicolumn{4}{c|}{\textbf{Pubmed}} \\ \hline
\textbf{K}                                                                                     & 5      & 10      & 15     & 20     & 5      & 10      & 15      & 20        \\ \hline
LOF                                                                             &   85.60     &    80.91     &    76.62    &  72.01      &    85.71    &    81.27 &    76.89     &     72.71        \\
OC-SVM                                                                          &   85.45     &    80.76     &    76.92    &  72.45      &  85.54      &    81.03     &    76.47     &     72.27        \\
Deep-SVDD                                                                             &   87.89     &    85.49     &    81.50    &  77.25      &      87.93&    84.63     &    80.96     &   77.07           \\
Radar                                                                                &   87.70     &    85.27     &    82.35    &   78.58     &     86.02   &  81.90       &    78.03     &     74.06  \\
GCN                                                                                &   85.23     &    83.38     &    76.70    &   71.05           &    86.95     &    83.57     &      80.11 &   76.41  \\

\textbf{SpecAE}                                          &       \textbf{94.31}          &    \textbf{90.81}    &    \textbf{86.67}     &   \textbf{82.05}    &   \textbf{94.98}     &   \textbf{90.50}     &    \textbf{85.96}     &      \textbf{81.52}               \\
\hline
\end{tabular}
\vspace{-5pt}
  \caption{Accuracy@K performance comparisons on two datasets. }
  \vspace{-8pt}
  \label{precision}
\end{table}

We evaluate the anomaly detection performance by comparing SpecAE with the five baseline methods.
Accuracy@K is utilized as the evaluation metric and the results are reported in Tab.~\ref{precision}. We output a ranking list based on the anomalous degree of different nodes, and we measure the effectiveness of every detection method in its top K ranked nodes. Then, we report the results in terms of ROC-AUC in Fig.~\ref{ROCCora}. For both metrics, SpecAE consistently achieves good performance. In addition, we make the following observations. (1) In general, the proposed SpecAE outperforms all of the baseline methods. Comparing with the methods which only utilize uni-modal information (nodal attributes in LOF, OC-SVM and Deep-SVDD), our proposed approach achieves better performance which validates the importance of applying heterogeneous multi-modal sources (nodal attributes and topological relations). 
(2) The performance of deep models (Deep-SVDD, GCN and SpecAE) are superior to the conventional anomaly detection methods in shallow structures  (LOF and OC-SVM). It verifies the effectiveness of extracting features to represent the nodal attributes on attributes networks in deep structures which can extract features in a more effective way. 
(3) SpecAE shows the importance of enlarging the model capacity by introducing the mutual interactions between nodal attributes with linkage connections and enabling topological relations to contain additional information. 
As can be observed, Spectral autoencoder can achieve better performance on anomaly detection application than pure GCN based model, which justifies the advantage of our proposed model that overcomes the drawbacks of the GCN, and adopts it into an anomaly detection scenario. We will quantitatively evaluate the specific properties of the SpecAE structure in the ablation analysis part.
\subsection{Ablation Studies}

\begin{table}[t]
    \centering
    \tiny
    \setlength{\tabcolsep}{8.8pt}
\begin{tabular}{|c|ccccc|}
 \hline
Metric & Accuracy & Precision & Recall     & F1     & AUC        \\ \hline
SpecAE-S                                                                          &   54.51     &    12.54 &    \textbf{59.63}   &  20.72      &  58.45              \\
SpecAE-N                                                                           &   90.51     &    52.40     &    52.59    &  52.50      &    73.65           \\
SpecAE-nr                                                                          &   82.24     &    11.07 &    11.11   &  11.09      & 50.53              \\
SpecAE $\alpha = 1$                                                                                  &   90.36     &    51.66     &  51.85      &   51.76     &    73.96        \\
SpecAE $\alpha = 0.7$                                                                            &   \textbf{91.93}     &    \textbf{54.98}     &    55.19    &  \textbf{55.08}      &      \textbf{77.15}         \\
\hline
\end{tabular}
\vspace{-4pt}
  \caption{Ablation and hyperparameter analysis.}
  \vspace{-6pt}
  \label{Ablation}
\end{table}

We conduct ablation studies to demonstrate the importance of individual components in SpecAE. We perform ablation analysis comparing SpecAE with four alternatives. (i) SpecAE-S: SpecAE without the representations for community anomaly detection, i.e., $\mathbf{Z}_{\mathcal{G}}$ and $\mathbf{Z}_{\mathcal{G}\text{error}}$; (ii) SpecAE-N: SpecAE without the representations for global anomaly detection, i.e., $\mathbf{Z}_{X}$ and $\mathbf{Z}_{X\text{error}}$; (iii)  SpecAE-nr: SpecAE without the reconstruction components $ \mathbf{Z}_{X\text{error}}$ and $\mathbf{Z}_{\mathcal{G}\text{error}}$; and (iv) training with an extreme value hyper-parameter $\alpha = 1$  (short-circuit the self features in the graph convolution and deconvolution). Results on Cora are shown in Tab.~\ref{Ablation}, indicating that the comprehensive method is superior to the variants, since it outperforms most of the ablation test settings where one component has been removed or replaced at a time from the full system and performed re-training and re-testing. Similar results could be observed on other datasets (Pubmed). Due to the reason of the space limit, we omit the detailed analysis. According to the results, we can find that the context information $\mathbf{Z}_{\mathcal{G}}$ provided by the SpecAE is crucial for anomaly detection in the attributed network (SpecAE-N, SpecAE outperform the other structures by $15\%$ in AUC). In the meanwhile, when we eliminate the $\mathbf{Z}_{X}$, the performance drops by $3.5\%$. Removing the reconstruction process hurts the performance by $27\%$,  which validates the reconstruction introduces useful information as indicators for anomaly detection.

\vspace{-4pt}

\subsection{A Case Study}
\begin{table}[t]
    \small
    \setlength{\tabcolsep}{4pt}
    \centering
\begin{tabular}{|c|c|}
\hline
\textbf{ (Anomaly).com}       & \textbf{Abnormal Keywords}  \\ \hline
\textbf{thismodernworld}       & siberians, fenno, scandinavia, \\& ultraviolet, primates, colder\\

\textbf{directorblue.blogspot}       & boiler, inflatable, choppers, streamline, \\ & rooftop, jettison, sheathed \\
\textbf{fafblog.blogspot}       & cheesecake, pies, crusts, barbecues, \\ & carrots, grapefruit pizzas \\
\textbf{balloon-juice}       & recapturing, wooly, buffy, \\ & talons, snarled, tails, dismemberment\\
\textbf{fringeblog}  & mosquitos, nerf, redwoods, mammoths, \\ & fungi, snail, hawked\\
\textbf{nomayo.mu.nu}       & pancakes, cheese, stewed, pork, brunch\\ \hline
\end{tabular}
\vspace{-3pt}
\caption{Keywords selected from the anomalous nodes on PolBlog.}
\vspace{-6pt}
\label{PolBlogTable}
\end{table}

We conduct a case study using the PolBlog dataset. The attributes of each node represent words appeared in each blog based on a given word dictionary. As shown in the Tab.~\ref{PolBlogTable}, six sampled anomalies  (nodes) are listed for more specific explanations. The abnormal keywords refer to the key information which appears in the anomaly bloggers, yet has low frequency and lacks political flavor in a global perspective among all bloggers. To be more specific, some of the posts actually focus on other fields like climates  (thismordernworld), food (nomayo.mu.nu, fafblog.blogspot), constructions and tools (directorblue.blogspot), huntings and creatures (ballon-juice, fringeblog). Comparing with the most popular political topics, these selected keywords in the abnormal bloggers are less relevance to the political events.
\vspace{-5pt}
\section{Conclusion and Future Work}
In this paper, we introduce an effective framework SpecAE for identifying anomalies in attributed networks. To detect global and community anomalies, we map the attributed network into two types of low-dimensional representations. The first type consists of node representations learned from an autoencoder and corresponding reconstruction errors. To learn the second type of representations, we design the novel graph deconvolution neural networks as the complementary operation to the graph convolution, aiming to reconstruct nodal attributes according to the topological relations. The two types of learned representations are applied to a Gaussian mixture model to perform anomaly detection. The tailored representation learning model and the GMM model are trained collaboratively. Experiential results demonstrate that SpecAE has superior performance over state-of-the-art algorithms on real-world datasets. Ablation analysis shows the effectiveness of each component of SpecAE. In future work, we will investigate involving attention models to give our anomaly detection framework an explainable result; developing robust frameworks through incorporating adversarial techniques.
\newpage
\bibliographystyle{named}
\bibliography{ijcai19}

\begin{thebibliography}{}

\bibitem[\protect\citeauthoryear{An and Cho}{2015}]{an2015variational}
Jinwon An and Sungzoon Cho.
\newblock Variational autoencoder based anomaly detection using reconstruction
  probability.
\newblock {\em Special Lecture on IE}, 2:1--18, 2015.

\bibitem[\protect\citeauthoryear{Breunig \bgroup \em et al.\egroup
  }{2000}]{breunig2000lof}
Markus~M Breunig, Hans-Peter Kriegel, Raymond~T Ng, and J{\"o}rg Sander.
\newblock Lof: identifying density-based local outliers.
\newblock In {\em ACM sigmod record}, volume~29, pages 93--104. ACM, 2000.

\bibitem[\protect\citeauthoryear{Chen \bgroup \em et al.\egroup
  }{2017}]{chen2017outlier}
Jinghui Chen, Saket Sathe, Charu Aggarwal, and Deepak Turaga.
\newblock Outlier detection with autoencoder ensembles.
\newblock In {\em Proceedings of the 2017 SIAM International Conference on Data
  Mining}, pages 90--98. SIAM, 2017.

\bibitem[\protect\citeauthoryear{Defferrard \bgroup \em et al.\egroup
  }{2016}]{Defferrard-etal16Convolutional}
Micha{\"e}l Defferrard, Xavier Bresson, and Pierre Vandergheynst.
\newblock Convolutional neural networks on graphs with fast localized spectral
  filtering.
\newblock In {\em NIPS}, pages 3844--3852, 2016.

\bibitem[\protect\citeauthoryear{Hawkins \bgroup \em et al.\egroup
  }{2002}]{hawkins2002outlier}
Simon Hawkins, Hongxing He, Graham Williams, and Rohan Baxter.
\newblock Outlier detection using replicator neural networks.
\newblock In {\em International Conference on Data Warehousing and Knowledge
  Discovery}, pages 170--180. Springer, 2002.

\bibitem[\protect\citeauthoryear{Huang \bgroup \em et al.\egroup
  }{2019}]{huang2019graph}
Xiao Huang, Qingquan Song, Yuening Li, and Xia Hu.
\newblock Graph recurrent networks with attributed random walks.
\newblock 2019.

\bibitem[\protect\citeauthoryear{Kipf and Welling}{2017}]{kipf2016semi}
Thomas~N Kipf and Max Welling.
\newblock Semi-supervised classification with graph convolutional networks.
\newblock {\em 5th International Conference on Learning Representations}, 2017.

\bibitem[\protect\citeauthoryear{Li \bgroup \em et al.\egroup
  }{2017}]{li2017radar}
Jundong Li, Harsh Dani, Xia Hu, and Huan Liu.
\newblock Radar: Residual analysis for anomaly detection in attributed
  networks.
\newblock {\em IJCAI’17}, 2017.

\bibitem[\protect\citeauthoryear{Li \bgroup \em et al.\egroup
  }{2018}]{li2018deeper}
Qimai Li, Zhichao Han, and Xiao-Ming Wu.
\newblock Deeper insights into graph convolutional networks for semi-supervised
  learning.
\newblock In {\em Thirty-Second AAAI Conference on Artificial Intelligence},
  2018.

\bibitem[\protect\citeauthoryear{Li \bgroup \em et al.\egroup
  }{2019}]{li2019deepstruc}
Yuening Li, Ninghao Liu, Jundong Li, Mengnan Du, and Xia Hu.
\newblock Deep structured cross-modal anomaly detection.
\newblock {\em IJCNN}, 2019.

\bibitem[\protect\citeauthoryear{Liu \bgroup \em et al.\egroup
  }{2017}]{liu2017accelerated}
Ninghao Liu, Xiao Huang, and Xia Hu.
\newblock Accelerated local anomaly detection via resolving attributed
  networks.
\newblock In {\em Proceedings of the 26th International Joint Conference on
  Artificial Intelligence}, pages 2337--2343. AAAI Press, 2017.

\bibitem[\protect\citeauthoryear{Liu \bgroup \em et al.\egroup
  }{2019}]{liu2019single}
Ninghao Liu, Qiaoyu Tan, Yuening Li, Hongxia Yang, Jingren Zhou, and Xia Hu.
\newblock Is a single vector enough? exploring node polysemy for network
  embedding.
\newblock {\em arXiv preprint arXiv:1905.10668}, 2019.

\bibitem[\protect\citeauthoryear{Ma \bgroup \em et al.\egroup
  }{2014}]{ma2014optimized}
Tinghuai Ma, Lu~Li, Sai Ji, Xin Wang, Yuan Tian, Abdullah Al-Dhelaan, and Mznah
  Al-Rodhaan.
\newblock Optimized laplacian image sharpening algorithm based on graphic
  processing unit.
\newblock {\em Physica A: Statistical Mechanics and its Applications},
  416:400--410, 2014.

\bibitem[\protect\citeauthoryear{McPherson \bgroup \em et al.\egroup
  }{2001}]{mcpherson2001birds}
Miller McPherson, Lynn Smith-Lovin, and James~M Cook.
\newblock Birds of a feather: Homophily in social networks.
\newblock {\em Annual review of sociology}, 27(1):415--444, 2001.

\bibitem[\protect\citeauthoryear{Perozzi \bgroup \em et al.\egroup
  }{2014}]{perozzi2014focused}
Bryan Perozzi, Leman Akoglu, Patricia Iglesias~S{\'a}nchez, and Emmanuel
  M{\"u}ller.
\newblock Focused clustering and outlier detection in large attributed graphs.
\newblock In {\em Proceedings of the 20th ACM SIGKDD international conference
  on Knowledge discovery and data mining}, pages 1346--1355. ACM, 2014.

\bibitem[\protect\citeauthoryear{Ruff \bgroup \em et al.\egroup
  }{2018}]{ruff2018deep}
Lukas Ruff, Nico G{\"o}rnitz, Lucas Deecke, Shoaib~Ahmed Siddiqui, Robert
  Vandermeulen, Alexander Binder, Emmanuel M{\"u}ller, and Marius Kloft.
\newblock Deep one-class classification.
\newblock In {\em International Conference on Machine Learning}, pages
  4390--4399, 2018.

\bibitem[\protect\citeauthoryear{Sakurada and
  Yairi}{2014}]{sakurada2014anomaly}
Mayu Sakurada and Takehisa Yairi.
\newblock Anomaly detection using autoencoders with nonlinear dimensionality
  reduction.
\newblock In {\em Proceedings of the MLSDA 2014 2nd Workshop on Machine
  Learning for Sensory Data Analysis}, page~4. ACM, 2014.

\bibitem[\protect\citeauthoryear{Sch{\"o}lkopf \bgroup \em et al.\egroup
  }{2000}]{scholkopf2000support}
Bernhard Sch{\"o}lkopf, Robert~C Williamson, Alex~J Smola, John Shawe-Taylor,
  and John~C Platt.
\newblock Support vector method for novelty detection.
\newblock In {\em Advances in neural information processing systems}, pages
  582--588, 2000.

\bibitem[\protect\citeauthoryear{Skillicorn}{2007}]{skillicorn2007detecting}
David~B Skillicorn.
\newblock Detecting anomalies in graphs.
\newblock In {\em Intelligence and Security Informatics, 2007 IEEE}, pages
  209--216. IEEE, 2007.

\bibitem[\protect\citeauthoryear{Song \bgroup \em et al.\egroup
  }{2007}]{song2007conditional}
Xiuyao Song, Mingxi Wu, Christopher Jermaine, and Sanjay Ranka.
\newblock Conditional anomaly detection.
\newblock {\em IEEE Transactions on Knowledge and Data Engineering},
  19(5):631--645, 2007.

\bibitem[\protect\citeauthoryear{Song \bgroup \em et al.\egroup
  }{2018}]{song2018deepmem}
Wei Song, Heng Yin, Chang Liu, and Dawn Song.
\newblock Deepmem: Learning graph neural network models for fast and robust
  memory forensic analysis.
\newblock In {\em Proceedings of the 2018 ACM SIGSAC Conference on Computer and
  Communications Security}, pages 606--618. ACM, 2018.

\bibitem[\protect\citeauthoryear{Yu \bgroup \em et al.\egroup
  }{2018}]{yu2018netwalk}
Wenchao Yu, Wei Cheng, Charu~C Aggarwal, Kai Zhang, Haifeng Chen, and Wei Wang.
\newblock Netwalk: A flexible deep embedding approach for anomaly detection in
  dynamic networks.
\newblock In {\em Proceedings of the 24th ACM SIGKDD International Conference
  on Knowledge Discovery \& Data Mining}, pages 2672--2681. ACM, 2018.

\bibitem[\protect\citeauthoryear{Zong \bgroup \em et al.\egroup
  }{2018}]{zong2018deep}
Bo~Zong, Qi~Song, Martin~Renqiang Min, Wei Cheng, Cristian Lumezanu, Daeki Cho,
  and Haifeng Chen.
\newblock Deep autoencoding gaussian mixture model for unsupervised anomaly
  detection.
\newblock 2018.

\end{thebibliography}

\end{document}